\documentclass{ecai}  

\usepackage{graphicx}
\usepackage{latexsym}

\usepackage{times}
\usepackage{soul}
\usepackage{url}
\usepackage[hidelinks]{hyperref}
\usepackage[utf8]{inputenc}
\usepackage[small]{caption}
\usepackage{amsmath}
\usepackage{amsthm}
\usepackage{booktabs}
\usepackage{algorithm}
\usepackage{algorithmic}
\usepackage{amssymb}
\usepackage{multirow}
\usepackage{enumitem}
\usepackage[table]{xcolor}
\usepackage{colortbl}
\usepackage[switch]{lineno}
\usepackage{float}
\usepackage{subfigure}

\def\ie{{\em i.e.}}
\def\eg{{\em e.g.}}
\def\etal{{\em et al.}}
\definecolor{igray}{gray}{.9}
\definecolor{iyellow}{RGB}{255,252,224}

\newcommand{\bcdot}{\boldsymbol{\cdot}}

\begin{document}

\begin{frontmatter}

\title{Semantic-Aware Dual Contrastive Learning for Multi-label Image Classification}


%


\author[A,D]{\fnms{Leilei}~\snm{Ma}}
\author[B,D]{\fnms{Dengdi}~\snm{Sun}}
\author[C]{\fnms{Lei}~\snm{Wang}}
\author[A,D,E]{\fnms{Haifeng}~\snm{Zhao}\thanks{Corresponding Author. Email: senith@163.com.}}
\author[A,D]{\fnms{Bin}~\snm{Luo}}


\address[A]{School of Computer Science and Technology, Anhui University, China}
\address[B]{School of Artificial Intelligence, Anhui University, China}
\address[C]{School of Computer Science and Engineering, Nanjing University of Science and Technology, China}
\address[D]{Anhui Provincial key Laboratory of Multimodal Cognitive Computing, Anhui University, China}
\address[E]{Institute of Artificial Intelligence, Hefei Comprehensive National Science Center, China}

%
%

\begin{abstract}
	Extracting image semantics effectively and assigning corresponding labels to multiple objects or attributes for natural images is challenging due to the complex scene contents and confusing label dependencies. Recent works have focused on modeling label relationships with graph and understanding object regions using class activation maps (CAM). However, these methods ignore the complex intra- and inter-category relationships among specific semantic features, and CAM is prone to generate noisy information. To this end, we propose a novel semantic-aware dual contrastive learning framework that incorporates sample-to-sample contrastive learning (SSCL) as well as prototype-to-sample contrastive learning (PSCL). Specifically, we leverage semantic-aware representation learning to extract category-related local discriminative features and construct category prototypes. Then based on SSCL, label-level visual representations of the same category are aggregated together, and features belonging to distinct categories are separated. Meanwhile, we construct a novel PSCL module to narrow the distance between positive samples and category prototypes and push negative samples away from the corresponding category prototypes. Finally, the discriminative label-level features related to the image content are accurately captured by the joint training of the above three parts. Experiments on five challenging large-scale public datasets demonstrate that our proposed method is effective and outperforms the state-of-the-art methods.  Code and supplementary materials are released on~\href{https://github.com/yu-gi-oh-leilei/SADCL}{https://github.com/yu-gi-oh-leilei/SADCL}.
\end{abstract}

\end{frontmatter}

\section{Introduction}
Multi-label image classification (MLIC) aims to assigning multiple labels to objects or attributes present in a natural image. As a fundamental task in computer vision, it is an essential component in many applications, such as attribute recognition~\cite{Yang2020CVPR}, weakly supervised semantic segmentation~\cite{Ru2022CVPR} and automatic image annotation~\cite{Jing2016Annotation}. In general, distinct from single-label classification, \textbf{\emph{the} MLIC \emph{task faces two main challenges}}: i) complex intra- and inter-category relationships which are hard to be modeled, and ii) various object scales, appearances, and layouts which make it challenging to extract image semantic information effectively.

\begin{figure}[t]
	\begin{center}
		\includegraphics[width=1.0\linewidth]{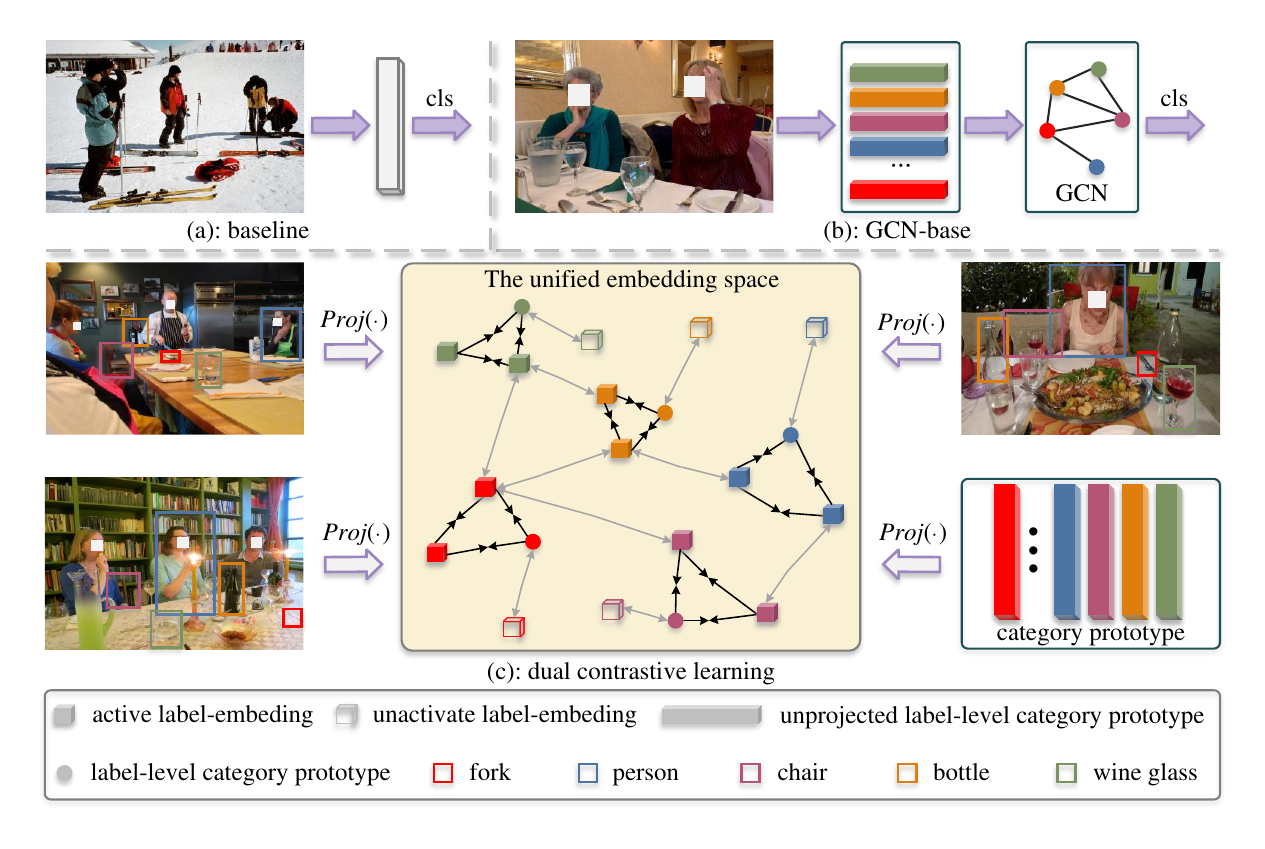}
		\vspace{-0.6cm}
		\caption{Motivation of the proposed dual contrastive learning. 
			(a) The baseline methods obtain a feature representation of multiple-category blends. 
			(b) GCN-based methods can easily cause overfitting.
			(c) With the intervention of category prototypes and contrastive learning, our method can obtain discriminative features.
		}
		\label{fig1}
	\end{center}
	\vspace{-0.6cm}
\end{figure}

For the first challenge, most existing works model the inter-category relationships based on the label co-occurrence dependency. One popular way~\cite{wang2016cnn,Wang2017ICCV} adopts Recurrent Neural Networks (RNN) or Long Short-Term Memory (LSTM) to model this relationship in a joint image-label embedding space, and then the co-occurred labels can be predicted sequentially.   Another mainstream research~\cite{chen2021learning,ye2020attention} focuses on utilizing Graph Convolutional Network (GCN) to capture label-wise relationships in semantic graphs statically or dynamically. Although promising results have been obtained, modeling label dependency is still difficult due to unreliable co-occurrence statistics~\cite{zhu2021residual}.

For the second challenge, region- and attention-based methods show their ability to discover objects with various scales and discontinuous regions. However, region-based approaches still have some drawbacks, \eg, high computational overhead~\cite{wu2021gm} and a large number of noisy region proposals~\cite{wei2015hcp}. 
Attention-based methods, such as You \etal~design cross-modal attention modules with cosine similarity to learning category-related regional features~\cite{you2020cma}.
Zhao \etal~develop a CAM-based module to generate category-specific activation maps which are further leveraged to convert cross-scale semantic feature maps into semantic-aware features~\cite{Zhao2021ICCV}. Nevertheless, the aforementioned methods have the following limitations: i) only inter-category relationships (intra-image) are considered, and intra-category relationships (cross-image) are ignored; ii) the localized semantic region or object region lacks discrimination.

To effectively address these limitations as well as the major challenges in multi-label image classification, we propose semantic-aware dual contrastive learning (SADCL) strategy, which is beyond image-level contrastive learning~\cite{khosla2020supervised} in building inter-category and intra-category discriminative correlation jointly. \textbf{\emph{To address the first challenge}},  our SADCL constructs contrastive representation learning from two aspects, \ie, sample-to-sample and prototype-to-sample, as shown in Figure~\ref{fig1}~(c). For the former, the semantic-aware representations, or label-level visual representations, from both intra- and cross-images are contrasted. Since multiple categories may occur concurrently in a multi-label image, conventional contrastive learning, which only considers intra-category contrastive representations from cross-image, is insufficient. Hence, we also add inter-category representations from multiple images for sample-to-sample contrastive learning, where modeling label correlations is involved. For the latter, we introduce visual prototypes and produce sample pairs from label-level visual representations and their corresponding prototypes. The visual prototype of one category preserves representative information of this category and can be viewed as a faithful category agent. The prototype-to-sample contrastive learning contributes to enlarging/decreasing the similarity between positive/negative label-level visual representations and their prototypes for each image. Note that we use the same project head for both contrastive learning, so the learned representations are in a unified embedding space. In addition, \textbf{\emph{to handle the second challenge}}, we model the context relationship among multiple objects and scenes at the front end of the framework, and generate an initial label-level visual representation with abundant semantic information through transformer autoencoder with multi-head attention mechanism. Then, the above dual contrastive learning is integrated to further optimize the label-level features in the unified embedding space, so as to obtain the more discriminative semantic-aware representations of an image.

Overall, in our method, discriminative label-level visual representations have the following characteristics:  i) they are derived from local discriminative regions of objects, ii) in the unified embedding space, negative label-level visual representations are more discriminative against associated category prototypes than positive ones, and iii) label-level visual representations are largely distinct between each category-pair with low co-occurrence, while similar between that with high co-occurrence.

Our contributions can be summarized in three-fold:

$\bullet$ We propose a novel Semantic-Aware Dual Contrastive Learning framework named SADCL for multi-label image classification, effectively learning more discriminative feature representation.

$\bullet$ Compared with class activation mapping (CAM), we leverage Semantic-Aware Representation Learning to accurately and easily locate the label-related image regions.

$\bullet$ Experiments on five challenging large-scale public datasets (MS-COCO, PASCAL VOC 2007\&2012, NUS-WIDE, and Visual Genome) show that our proposed method is effective and outperforms the state-of-the-art methods.

\vspace{-0.1cm}
\section{Related Works}
\textbf{Multi-Label Image Classification.} With the development of computer vision, multi-label image
classification (MLIC) task has received extensive attention and made remarkable progress. The related researches could be roughly divided into two aspects \ie{}, relation-based methods and region/attention-based methods.

Since the relationship of multiple objects co-occurring in a multi-label image can be meaningful for classification, modeling label correlation has become a hot topic in MLIC.	
Pioneer works~\cite{wang2016cnn,Wang2017ICCV} utilize RNN or LSTM to transform labels and an image into a joint embedding space and predict labels in a pre-defined orderly way.
Despite this, RNN-based methods only model label-local relations while ignoring label-global relations.
To address this limitation, GCN-based methods propagate information between nodes over a graph and explore labels-wise relationships.
For example, Li~\etal~\cite{li2020learning} design an adaptive label graph learning module with sparse correlation constraints to reduce the hassle of hand-crafted graphs.
To overcome the dependence of GCN on learning only pair-wise labels, Wu~\etal~\cite{wu2020adahgnn} introduce adaptive hypergraph neural network to model the higher-order semantic relationships among labels automatically.
On the other hand, region and attention-based methods have been handling various object scales and appearances in MLIC tasks.
Many existing works~\cite{wu2021gm} adopt object detection methods to generate a set of semantic-aware instances and construct semantic label graphs.
Nevertheless, object detection methods require the pre-trained detector with additional bounding box annotations.
Exploiting attention mechanism to extract salient object regions is another popular fashion.
For example, Ye~\etal~\cite{ye2020attention} design a semantic attention module to generate category-specific activation maps, and then obtain content-aware category representations.
Zhao~\etal~\cite{Zhao2021ICCV} propose a cross-attention module to suppress the noise between different scales and enhance the structural information of small objects.
Liu~\etal~\cite{liu2021q2l} make use of label embeddings as queries to learn class-specific representations via cross-modal attention.

\paragraph{Contrastive learning.} Contrastive learning is an effective method to strengthen the discrimination of the leaned representations, which aims to bring the samples in positive sample pairs closer together in a unified embedding space while pushing samples in negative pairs away.
For example, He~\etal~\cite{he2020momentum} build a dynamic dictionary with a queue and a moving average encoder for unsupervised contrastive learning. 
Khosl~\etal~\cite{khosla2020supervised} leverage label information for supervised contrastive learning, which sets samples with the same label as positive pairs while samples with different labels as negative pairs.
In addition, Li~\etal~\cite{li2021prototypical} proposed prototypical contrastive learning for unsupervised presentation learning, where prototypes represent class-agnostic semantic information, but we propose prototypes for class-specific semantic representations.
Although contrastive learning for MLIC has been explored preliminarily by Dao~\etal~\cite{dao2021multi}, it fails to model label correlation. Moreover, two-time augmentation of an image makes it inefficient.

\vspace{-0.1cm}
\section{Methodology}
\begin{figure*}[ht]
	\centering
	\includegraphics[scale=0.67]{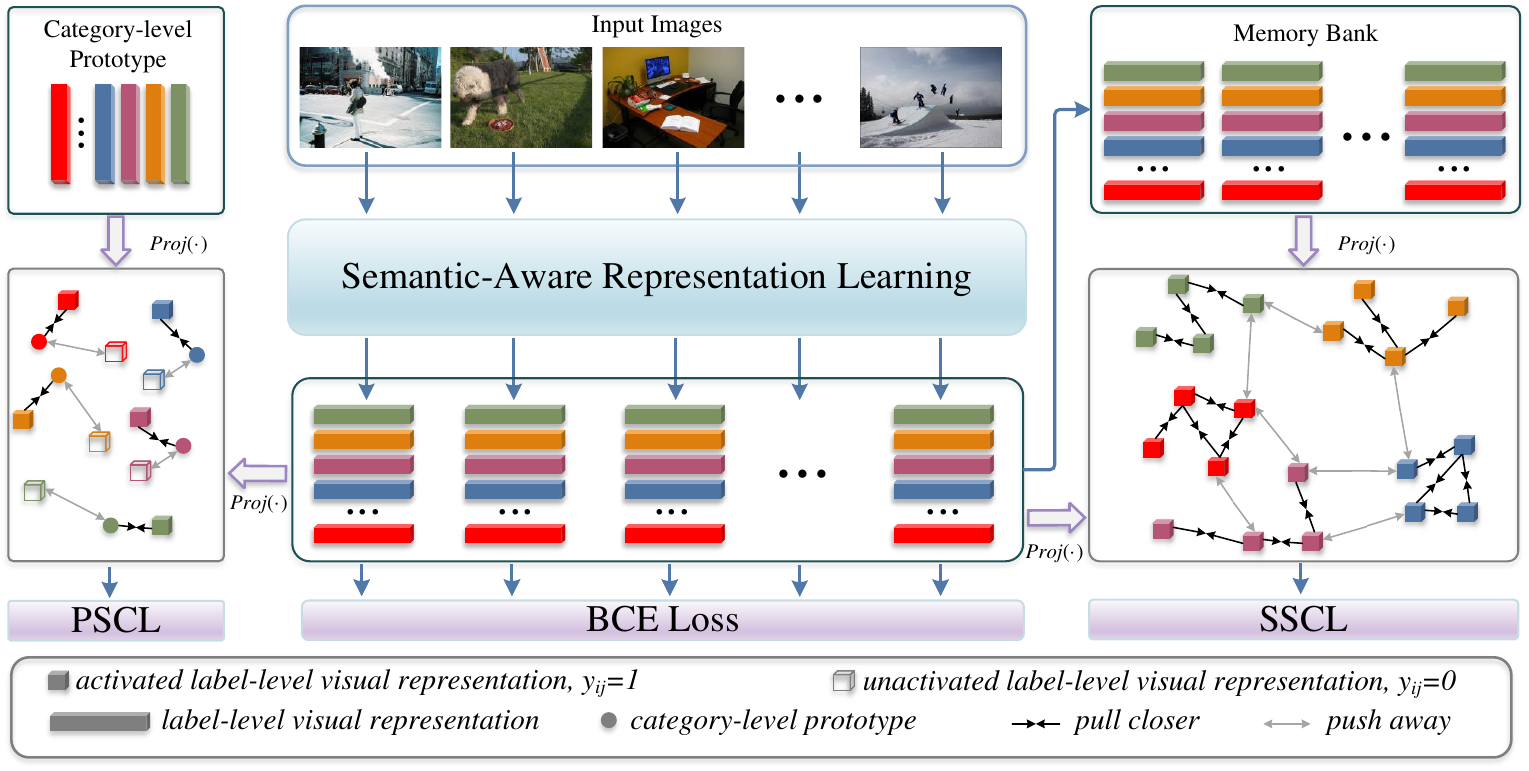}
	\caption{The overview of the SADCL framework. \textbf{Note that a solid square represents an active state, and a hollow square represents an inactive state}. Meanwhile, different colors represent different categories. \text{Proj}($\cdot$) is a projection network.}
	
	\label{fig2}
\end{figure*}
\subsection{Overview and Preliminary}
In an MLIC task, given a minibatch of input images $\mathcal{D}=\{(I_i,Y_i)\}_{i=1}^N$ with $L$ categories, where $N$ is the batch size, our goal is to build a visual model $F(\cdot)$ to predict the label ${Y}_i$ of the given image $I_i$. If an image $I_i \in \mathcal{D}$ contains the $j$-th category, the associated label $y_{ij} \in Y_i$ will be $1$, and vice versa.

For this purpose, we propose a Semantic-Aware Dual Contrast Learning (SADCL) framework that learns more discriminative representations for MLIC task. As shown in Figure~\ref{fig2}, the architecture of the proposed SADCL consists of three components: Semantic-Aware Representation Learning (SARL), Sample-to-Sample Contrastive Learning (SSCL), and Prototype-to-Sample Contrastive Learning (PSCL). SARL decomposes the image into $L$ label-level visual representations (features), and then SSCL and PSCL optimize the distribution of label-level representations from the perspective of samples and category prototype respectively, so that in a unified embedded space the homogeneous and activated features tend to aggregate, while heterogeneous or unactivated features are pushed away. Finally, the optimized label-level representation is fed into a classifier to predict whether each category is present or not. In the inferring stage, no more contrastive learning is needed, and we use SARL to directly obtain label-level visual representation for classification.

\subsection{Semantic-Aware Representation Learning}\label{SARL}
\begin{figure}[!t]
	\begin{center}
		\includegraphics[width=1.0\linewidth]{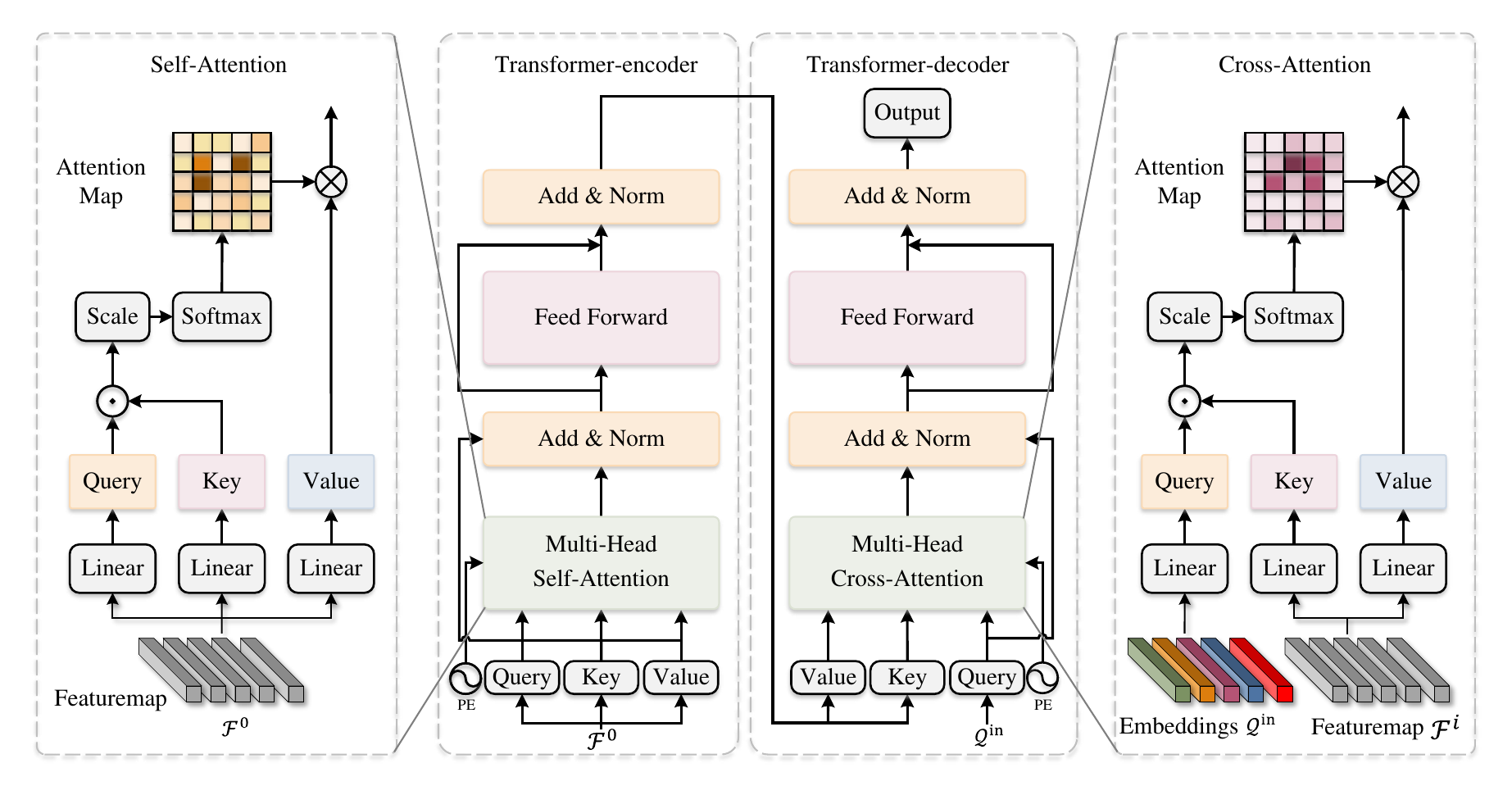}
		\caption{Illustrations of Transformer-encoder and Transformer-decoder.
		}\label{fig3}
	\end{center}
\end{figure}

To accurately extract semantic information from complex image content, following~\cite{dao2021multi,liu2021q2l}, we first model the contextual spatial layout among multiple objects and scene to generate feature maps with abundant contextual semantic information. Specifically, the initial spatial feature maps $\mathcal{F}\in \mathbb{R}^{H_0\times W_0\times d}$ of the input image are extracted through CNN, where $H_0$, $W_0$ and $d$ are the length, width, and channel number, and reshaped as a sequence feature form: $\mathcal{F }^0\in \mathbb{R}^{H_0 W_0\times d}$. Then the sequential features $\mathcal{F}^0$ and the corresponding positional embedding $PE$ are input to a transformer-encoder, as shown on the left half of Figure ~\ref{fig3}. In this way, the output features with rich contextual semantic information will be generated as follows:
\begin{equation}
	\mathcal{F}^i = \text{FFN} (\text{SelfAttBlock}(\widetilde{\mathcal{F}}^{i-1}, PE))~,
	\label{eq:encoder}
\end{equation}
where multi-head self-attention (SelfAttBlock) and feed-forward neural network (FFN) are the main components of the transformer-encoder. Eq. \ref{eq:encoder} is a layerwise iterative form, and here we can simply take the output features to be $\mathcal{F}^1$.

After that, as shown on the right half of Figure~\ref{fig3}, we utilize the learnable category semantic embeddings as queries $\mathcal{Q}^{in}\in \mathbb{R}^{C\times d}$, and the spatial features $\mathcal{F}^i$ as keys and values, and perform a transformer-decoder to generate the label-level visual representations as follows. 
\begin{equation}
	\mathcal{Q}^i = \text{FFN} (\text{CrossAttBlock}(\widetilde{\mathcal{F}}^{i-1}, \widetilde{\mathcal{Q}}^{i-1}, PE))~,
\end{equation}
where multi-head cross-attention (CrossAttBlock) locates category-related discriminative regions, and different heads focus on information from different parts of the same object. Consequently, the regions in the feature maps with high correlation to $\mathcal{Q}^{i-1}$ are aggregated in $\mathcal{Q}^{i}$ and updated layer-by-layer. In this way, the label-related regions can be located as accurately as possible to extract semantic-aware visual representations. For example, if an image contains a \textit{cat}, the region in the image from the \textit{cat} will be highlighted and have a higher correlation with the corresponding category semantic embedding.
\subsection{Projection Network for Contrastive Learning}
\label{PNCL}
After caputing the initial label-level visual representations, perform a projection transformation and embed the learned features into a unified vector space for dual contrastive learning. As mentioned above,
given an image $I_i$, the label-level features $\mathcal{Q}^{out}$ is generated by backbone and semantic-aware representation learning. Then, $\mathcal{Q}^{out}$ is mapped to a unified embedding space using a projection network, which consists of two linear layers and an activation function:
\begin{equation}
	x_{i} = \text{Proj}(\mathcal{Q}^{out}) \in \mathbb{R}^{L\times d}~,
\end{equation}
where $x_{i}$ is the projected label-level visual representation for the image $I_i$.
Since contrastive learning is conducted in the projected space, we can also refer the unified embedding space to the projected space.

\subsection{Sample-to-Sample Contrastive Learning}
\label{SSCL}
In the single-label image classification task, most previous contrastive learning methods~\cite{he2020momentum,khosla2020supervised} are based on image-level contrastive losses.
These methods define positive and negative sample pairs as follows: in a mini-batch of the input images, the image-level representations with different augmentations belonging to the same category are defined as positive sample pairs, and the image-level representations from other categories or images are defined as negative sample pairs.
However, for multi-label image classification, an image usually contains multiple categories, making it challenging to define positive or negative sample pairs at the image level.
Instead of using image-level representations like existing works on contrastive learning, we propose sample-to-sample contrastive learning, which defines positive and negative sample pairs at the label level.
In the minibatch of $N$ input images, the projected label-level visual representations from the memory bank and SADCL are aggregated into a set $\mathcal{X}=\{(x_{ij},y_{ij})\ \vert i \in \{1, \cdots, 2N\}; j \in \{1, \cdots, L\}$, where $x_{ij} \in \mathbb{R}^{d}$ is the label-level visual representation about the $j$-th category in the $i$-th image, $y_{ij} \in \{0,1\}$ is the ground-truth associated with $x_{ij}$.
Hereafter, we view $x_{ij}$ as an activated label-level visual representation, when $y_{ij}=1$, vice versa.
After that, we define positive and negative sample pairs in the activated label-level visual representation set $\mathcal{I} = \{x_{ij} \in \mathcal{X} | y_{ij} = 1\}$.
Here, the $x_{ij} \in \mathcal{I}$ is called the anchor, and the $x_{ij(p)}$ is referred to as the positive.
Note that the positive set is formulated in the following form $x_{ij(p)} \in \mathcal{P}_{ss}(i,j) = {\{x_{kj} \in \mathcal{A}(i,j)| y_{kj} = y_{ij} = 1\}}$, where $\mathcal{A}(i,j) = \mathcal{I} \setminus x_{ij}$ is the set including $\mathcal{I}$ and excluding $x_{ij}$.
In summary, the sample-to-sample contrastive loss for the anchor $x_{ij}$ are defined as follows:
\begin{equation}
	\begin{split}
		\mathcal{L}_{S2S}^{ij}&=\sum\limits_{x_{ij(p)} \!\in \mathcal{P}_{ss}(i,j)}\!
		\log\frac{\exp(x_{ij} \! \bcdot \! x_{ij(p)}/\tau)}{\sum_{x_a \! \in \mathcal{A}(i,j)} \! \exp(x_{ij} \! \bcdot \! x_a/\tau))}~,\\
		\mathcal{L}_{S2S} &= \sum_{x_{ij} \in \mathcal{I}} \frac{-1}{|\mathcal{P}_{ss}(i,j)|} \mathcal{L}_{S2S}^{ij}~,
	\end{split}
\end{equation}
where $\tau \in \mathcal{R}^{+}$ denotes a temperature parameter.

In addition, we propose a memory bank module because an image needs to be augmented twice in the conventional contrastive learning approach, which will double the image encoding calculation.
The memory bank stores activated label-level visual representations ($y_{i,j}\!=\!1$) from each training iteration, replacing the original representations.

\subsection{Prototype-to-Sample Contrastive Learning}
\label{PSCL}
SSCL only considers using activated label-level visual representations ($y_{ij}\!=\!1$) to perform contrastive learning. Furthermore, an image contains only a small number of categories, resulting in many label-level visual representations that are not active ($y_{ij}\!=\!0$).
This leads to inactive label-level visual representations that do not participate in contrastive learning, thereby restricting the model from obtaining more discriminative representations.
To alleviate this problem, we design a prototype-based contrastive learning loss to mine this unactivated label-level visual representation information entirely. As shown in Figure~\ref{fig2}, in the projected space, the activated label-level visual representations are as close as possible to the corresponding category prototypes. In contrast, inactive label-level visual representations are as far away as possible from the center of the related category prototypes. 

In line with this goal, we propose a category-level prototype module, which can be regarded as the center of the category.
The label-level visual representations produced during model training are averaged to obtain category-level prototypes $\mathcal{C}^{in} $. 
To ensure that the label-level visual representation is reliable, we propose a screening scheme: $\mathcal{Q}^{out}$ is the activation state ($y_{ij}=1$), and the corresponding prediction score is greater than a fixed threshold $\epsilon$. 
The process to obtain the category-level prototype is formulated as follows:
\begin{equation}
	\label{eq:prototype}
	\begin{split}
		\mathcal{C}^{in}_j &= \frac{1}{N_j} \sum\limits_{i=1}^{N}\mathcal{Q}^{out}_{ij}\cdot \mathbb{I}(\text{pred}(\mathcal{Q}^{out}_{ij}) \geq \epsilon)\cdot \mathbb{I}(y_{ij}=1),\\ \mathcal{C}^{in} &= [\mathcal{C}^{in}_1,\ldots,\mathcal{C}^{in}_j,\ldots ,\mathcal{C}^{in}_L]^\mathrm{T}\in\mathbb{R}^{L\times d}~,
	\end{split}
\end{equation}
where ${N_j}$ is the number of label-level visual representations $\mathcal{Q}^{out}_{ij}$ related to category $j$.
As a result of mapping $\mathcal{C}^{in}$ to the projection space, $\mathcal{C}^{out}\in \mathbb{R}^{L\times d}$ is obtained for contrastive learning by using the projection network.
For a specific category $j$, if we define $p_j$ as an anchor, the label-level visual representation for the category $j$: the activated $x_{ij(p)}$ as the set of positive set $\mathcal{P}_{sp}(i,j)$ and the inactivated $x_{ij}$ as the set of negative set $\mathcal{N}_{sp}(i,j)$.

Based on these analyses, the prototype-to-sample contrastive loss is designed as:

\begin{align}
	\label{eq:pscl}
	\mathcal{L}_{P2S} = -\sum\limits_{j \in C} \log \frac{\sum\limits_{x_{ij(p)} \in \mathcal{P}_{sp}(i,j)} \exp( c_j \bcdot x_{ij(p)}/\tau)}{\sum\limits_{x_{ij} \in \mathcal{P}_{sp}(i,j) \cup \mathcal{N}_{sp}(i,j)  }  \exp( c_j \bcdot x_{ij}/\tau)}
\end{align}

\subsection{Classification and Optimizing}
With the label-level visual representations $\mathcal{Q}^{out}$ obtained by SARL, we treat each $\mathcal{Q}^{out}$ prediction as a binary classification task and design a group of binary classifiers, consisting of a linear transformation layer with sum operation and a nonlinear activation function:
\begin{equation}
	s_l =  \sigma \left( \text{sum}\left(W_l^T \mathcal{Q}^{out}_{l}+b_l \right) \right)~,
	\label{eq:cls}
\end{equation}
where $W_l^T \in\mathbb{R}^{d}$ is a learnable weight for category $l$, $b_l\in\mathbb{R}^1$ is a bias, $\sigma$ is sigmoid activation, 
and $s_l$ denotes the prediction score.

Following conventional multi-label image classification works, we adopt BCE loss as the classification loss:
\begin{equation}
	\label{eq:bce}
	\mathcal{L}_{BCE} = \sum_{j=1}^L y_{ij} \log s_{ij} + (1-y_{ij})\text{log}(1 - s_{ij})~,
\end{equation}
where $s_{ij}$ is the score for predicting $x_{ij}$, and $y_{ij}$ is the corresponding ground-truth label.
As a final step, classification loss, sample-to-sample contrastive loss and prototype-to-sample contrastive learning jointly optimize our model:
\begin{equation}
	\label{eq:loss}
	\mathcal{L} = \mathcal{L}_{S2S} + \mathcal{L}_{P2S} + \mathcal{L}_{BCE}~.
\end{equation}

\begin{table*}[!ht]
	\centering{
	\renewcommand{\arraystretch}{1.1}
	\resizebox{\textwidth}{!}{
		\begin{tabular}{c|c|c| cccccc | cccccc }		 	 
			\hline
			\multirow{2}{*}{Methods} & \multirow{2}{*}{($R_{train}, R_{test}$)} & \multirow{2}{*}{mAP} & \multicolumn{6}{c|}{All}  & \multicolumn{6}{c}{Top 3}		
			\\
			&                                              &      & CP   & CR   & CF1  & OP   & OR   & OF1  & CP   & CR   & CF1  & OP   & OR   & OF1  \\ \hline \hline
			CNN-RNN~\cite{wang2016cnn}      &$(-,-)$      & 61.2 & -    & -    & -    & -    & -    & -    & 66.0 & 55.6 & 60.4 & 69.2 & 66.4 & 67.8 \\
			RNN-Att~\cite{Wang2017ICCV}&$(-,-)$      & -    & -    & -    & -    & -    & -    & -    & 79.1 & 58.7 & 67.4 & 84.0 & 63.0 & 72.0 \\
			ResNet101\textbf{$^*$}\cite{He2016CVPR}     &$(448,448)$  & 81.5 & 82.1 & 71.2 & 76.0 & 84.6 & 75.4 & 79.7 & 85.9 & 62.9 & 71.6 & 89.6 & 66.1 & 76.1 \\	
			MLGCN~\cite{chen2019multi}      &$(448,448)$  & 83.0 & 85.1 & 72.0 & 78.0 & 85.8 & 75.4 & 80.3 & 89.2 & 64.1 & 74.6 & 90.5 & 66.5 & 76.7 \\
			MS-CMA~\cite{you2020cma}        &$(448,448)$  & 83.8 & 82.9 & 74.4 & 78.4 & 84.4 & 77.9 & 81.0 & 88.2 & 65.0 & 74.9 & 90.2 & 67.4 & 77.1 \\
			P-GCN~\cite{chen2021learning}    &$(448,448)$  & 83.2 & 84.9 & 72.7 & 78.3 & 85.0 & 76.4 & 80.5 & 89.2 & 64.3 & 74.8 & 90.0 & 66.8 & 76.7 \\
			GM-MLIC~\cite{wu2021gm}         &$(448,448)$  & 84.3 & 87.3 & 70.8 & 78.3 & 88.6 & 74.8 & 80.6 & 90.6 & 67.3 & 74.9 & 94.0 & 69.8 & 77.8 \\	
			MCAR~\cite{gao2021learning}      &$(448,448)$  & 83.8 & 85.0 & 72.1 & 78.0 & 88.0 & 73.9 & 80.3 & 88.1 & 65.5 & 75.1 & 91.0 & 66.3 & 76.7 \\
			TDRG~\cite{Zhao2021ICCV}         &$(448,448)$  & 84.6 & 86.0 & 73.1 & 79.0 & 86.6 & 76.4 & 81.2 & 89.9 & 64.4 & 75.0 & 91.2 & 67.0 & 77.2 \\
			CCD-R101~\cite{liu2022contextual}&$(448,448)$  & 84.0 & 87.2 & 70.9 & 77.3 & 88.8 & 74.6 & 81.1 & 89.7 & 63.9 & 72.9 & 92.0 & 66.5 & 77.2 \\
			MulCon~\cite{dao2021multi}       &$(448,448)$  & 84.9 & 84.0 & 74.8 & 79.2 & 85.6 & 78.0 & 81.6 & 87.8 & 65.9 & 75.3 & 90.5 & 67.9 & 77.6 \\ 
			Query2Label~\cite{liu2021q2l}    &$(448,448)$  & 84.9 & 84.8 & 74.5 & 79.3 & 86.6 & 76.9 & 81.5 & 78.0 & 69.1 & 73.3 & 80.7 & 70.8 & 75.4 \\
			CPSD~\cite{huang2022cpsd}        &$(448,448)$  & 84.9 & 88.4 & 71.7 & 79.2 & 89.3 & 74.8 & 81.4 & -    & -    & -    & -    & -    & -    \\
			\hline
			\rowcolor{iyellow}Ours(SADCL)      &$(448,448)$  & 85.6 & 84.6 & 76.0 & 79.8 & 86.0 & 78.5 & 82.1 & 88.9 & 66.6 & 74.9 & 91.0 & 68.3 & 78.0 \\	\hline \hline
			ADDGCN~\cite{ye2020attention}    &$(448,576)$  & 85.2 & 84.7 & 75.9 & 80.1 & 84.9 & 79.4 & 82.0 & 88.8 & 66.2 & 75.8 & 90.3 & 68.5 & 77.9 \\
			SSGRL~\cite{Chen2019ICCV}        &$(576,576)$  & 83.8 & 89.9 & 68.5 & 76.8 & 91.3 & 70.8 & 79.7 & 91.9 & 62.5 & 72.7 & 93.8 & 64.1 & 76.2 \\
			AdaHGNN~\cite{wu2020adahgnn}     &$(576,576)$  & 85.0 &  -   & -    & 79.9 &  -   &   -  & 81.8 &   -  &  -   & 75.5 &  -   &  -   & 77.6 \\
			C-Tran~\cite{Tianlu2021Ctran}    &$(576,576)$  & 85.1 & 86.3 & 74.3 & 79.9 & 87.7 & 76.5 & 81.7 & 90.1 & 65.7 & 76.0 & 92.1 & 71.4 & 77.6 \\
			TDRG~\cite{Zhao2021ICCV}         &$(576,576)$  & 86.0 & 87.0 & 74.7 & 80.4 & 87.5 & 77.9 & 82.4 & 90.7 & 65.6 & 76.2 & 91.9 & 68.0 & 78.1 \\
			CCD-R101~\cite{liu2022contextual}&$(576,576)$  & 85.3 & 88.3 & 73.1 & 80.2 & 88.8 & 76.3 & 82.1 & 91.0 & 65.2 & 76.0 & 92.3 & 67.3 & 77.9 \\    \hline 
			\rowcolor{iyellow}Ours(SADCL)      &$(448,576)$  & 86.8 & 86.4 & 77.0 & 81.1 & 87.7 & 79.1 & 83.2 & 90.0 & 67.4 & 75.7 & 92.0 & 68.7 & 78.7 \\	\hline
			
		\end{tabular}
		}
		\caption{Comparisons with state-of-the-art methods on the MS-COCO dataset.\emph{*} indicates the reproduced results of our implementation. All metrics are in \%.}
		\label{table: COCO benchmark}
	}
\end{table*}
\section{Experiments}
\subsection{Experimental Setup}
\paragraph{Datasets.}
To evaluate the proposed method, we used the five most popular benchmark datasets: MS-COCO~\cite{lin2014microsoft}, NUS-WIDE~\cite{chua2009nus}, Visual Genome~\cite{krishna2017visual}, PASCAL VOC 2007~\cite{everingham2015pascal}, and PASCAL VOC 2012~\cite{everingham2015pascal}. 
All data collections are natural image collections containing categories commonly found in natural scenes.
The details of each dataset are shown in Table~\ref{tab:dataset}.
\begin{table}
	\centering
	\begin{tabular}{c|cccc}
		\hline
		Dataset        & \#Train  & \#Test  & \#Category  & \#Label Cardinality  \\
		\hline
		MS-COCO         & 82783    & 40504  & 80  & 2.9 \\
		NUS-WIDE        & 125449   & 83898  & 81  & 2.4  \\
		VOC2007         & 5011     & 4952   & 20  & 1.5 \\
		VOC2012         & 11540    & 10991  & 20  & 1.4 \\
		VG500           & 82904    & 10000  & 500 & 13.6  \\
		\hline
	\end{tabular}
	\caption{Statistics for the popular multi-label image classification dataset, including image numbers, category numbers, and label cardinality. Note that Visual Genome is referred to as VG500, PASCAL VOC 2007 is called VOC2007, and PASCAL VOC 2012 is named VOC2012.}
	\label{tab:dataset}
	\vspace{-0.5cm}
\end{table}

\paragraph{Evaluation Metrics.}
To evaluate our method, we adopt the average precision (AP) per category and mean of average precision (mAP) across all categories for evaluation.
As the three datasets (MS-COCO, NUS-WIDE, Visual Genome) contain so many categories, it is impossible to present the AP of each category, so in addition to mAP, we also show the following evaluation metrics: overall precision (OP), recall (OR), F1-measure (OF1) and per-category precision (CP), recall (CR), and F1-measure (CF1), computed over all prediction scores and top-3 highest prediction scores. All metrics are in \%. 

\paragraph{Implementation Details.}
For fair comparisons, ResNet-101~\cite{He2016CVPR} pre-trained on ImageNet-1k is adopted as our backbone. 
To accelerate model convergence, we use AdamW as the optimizer, and OneCycleLR is used as the learning rate scheduler, with the maximum learning rate being $5e\!\!-\!\!5$ under a batch size of 64.
For data augmentation and regularization, we employ the tricks suggested in ~\cite{liu2021q2l}: RandAugment, Cutout with a factor of 0.5, weight decay of $5e\!-\!3$.
Our method can converge fast, which achieves optimal results in less than 20 epochs.
\subsection{Comparison with State-of-The-Art Methods}
\paragraph{Results on MS-COCO.}
MS-COCO is the most widely used large-scale dataset with challenges for multi-label image classification. It has the following characteristics: common categories, various object scales, and imbalanced classes.
As shown in Table ~\ref{table: COCO benchmark}, compared with state-of-the-art methods, our method substantially outperforms other methods in all evaluation metrics, especially on mAP, CF1, and OF1.
Most methods use 448$\times$448 image resolution as input during the training and testing stage, but some works change the input image resolution to 576$\times$576.
To provide a fair comparison, we report the results only after modifying the input resolution to 576$\times$576 during the testing phase.

\paragraph{Results on PASCAL VOC 2007 and 2012.}
PASCAL VOC 2007 and 2012 are used as datasets for object detection and segmentation and evaluated for the MLIC task.
Since these two datasets cover 20 common categories, we report AP per category in addition to mAP.
As shown in Table~\ref{table: VOC2007 benchmark}, in terms of mAP, our proposed method achieves 96.4\%, outperforming other methods with a larger margin.
In addtion, the increase in the number of images trained by VOC2012 futher influences the results. 
As shown in Table~\ref{table: VOC2012 benchmark}, our method outperforms existing methods on VOC2012 with a larger margin.
Note that our method also outperforms other methods in small object recognition, \eg{}~\textit{bird},~\textit{cat}, and~\textit{plant}.

\begin{table*}[ht]
	\centering{
		 \setlength\tabcolsep{3pt}
		\resizebox{\textwidth}{!}{
			\begin{tabular}{c|c|c|c|c|c|c|c|c|c|c|c|c|c|c|c|c|c|c|c|c||c}
				\hline
				Methods & aero & bike & bird & boat &
				bottle & bus & car & cat & chair & 
				cow & table & dog & horse & motor & 
				person & plant & sheep & sofa & train & tv &
				mAP \\ \hline \hline
				CNN-RNN~\cite{wang2016cnn} & 96.7 & 83.1 & 94.2 & 92.8 & 61.2 & 82.1 & 89.1 & 94.2 & 64.2 & 83.6 & 70.0 & 92.4 & 91.7 & 84.2 & 93.7 & 59.8 & 93.2 & 75.3 & 99.7 & 78.6 & 84.0 \\
				ResNet-101\textbf{$^*$}~\cite{He2016CVPR}            & 99.0 & 98.4 & 97.5 & 96.0 & 81.4 & 97.3 & 97.3 & 97.1 & 79.6 & 96.0 & 88.1 & 97.5 & 98.5 & 95.8 & 98.8 & 85.9 & 97.2 & 84.6 & 98.8 & 92.0 & 93.8 \\
				RNN-Att~\cite{Wang2017ICCV}        & 98.6 & 97.4 & 96.3 & 96.2 & 75.2 & 92.4 & 96.5 & 97.1 & 76.5 & 92.0 & 87.7 & 96.8 & 97.5 & 93.8 & 98.5 & 81.6 & 93.7 & 82.8 & 98.6 & 89.3 & 91.9 \\
				SSGRL$^\star$~\cite{Chen2019ICCV}                & 99.5 & 97.1 & 97.6 & 97.8 & 82.6 & 94.8 & 96.7 & 98.1 & 78.0 & 97.0 & 85.6 & 97.8 & 98.3 & 96.4 & 98.1 & 84.9 & 96.5 & 79.8 & 98.4 & 92.8 & 93.4 \\
				ML-GCN~\cite{chen2019multi}              & 99.5 & 98.5 & 98.6 & 98.1 & 80.8 & 94.6 & 97.2 & 98.2 & 82.3 & 95.7 & 86.4 & 98.2 & 98.4 & 96.7 & 99.0 & 84.7 & 96.7 & 84.3 & 98.9 & 93.7 & 94.0 \\
				P-GCN~\cite{chen2021learning}            & 99.6 & 98.6 & 98.4 & 98.7 & 81.5 & 94.8 & 97.6 & 98.2 & 83.1 & 96.0 & 87.1 & 98.3 & 98.5 & 96.3 & 99.1 & 87.3 & 95.5 & 85.4 & 98.9 & 93.6 & 94.3 \\
				ADDGCN$^\star$~\cite{ye2020attention}           & 99.8 & 99.0 & 98.4 & 99.0 & 86.7 & 98.1 & 98.5 & 98.3 & 85.8 & 98.3 & 88.9 & 98.8 & 99.0 & 97.4 & 99.2 & 88.3 & 98.7 & 90.7 & 99.5 & 97.0 & 96.0 \\
				KGGR$^\star$~\cite{chen2022knowledge}	     	  & 99.3 & 98.6 & 97.9 & 98.4 & 86.2 & 97.0 & 98.0 & 99.2 & 82.6 & 98.3 & 87.5 & 99.0 & 98.9 & 97.4 & 99.1 & 86.9 & 98.2 & 84.1 & 99.0 & 95.0 & 95.0 \\
				DSDL~\cite{zhou2021dsdl}                 & 99.8 & 98.7 & 98.4 & 97.9 & 81.9 & 95.4 & 97.6 & 98.3 & 83.3 & 95.0 & 88.6 & 98.0 & 97.9 & 95.8 & 99.0 & 86.6 & 95.9 & 86.4 & 98.6 & 94.4 & 94.4 \\
				GM-MLIC~\cite{wu2021gm}                  & 99.4 & 98.7 & 98.5 & 97.6 & 86.3 & 97.1 & 98.0 & 99.4 & 82.5 & 98.1 & 87.7 & 99.2 & 98.9 & 97.5 & 99.3 & 87.0 & 98.3 & 86.5 & 99.1 & 94.9 & 94.7 \\
				TDRG~\cite{Zhao2021ICCV}                 & 99.9 & 98.9 & 98.4 & 98.7 & 81.9 & 95.8 & 97.8 & 98.0 & 85.2 & 95.6 & 89.5 & 98.8 & 98.6 & 97.1 & 99.1 & 86.2 & 97.7 & 87.2 & 99.1 & 95.3 & 95.0 \\
				MulCon$^\star$~\cite{dao2021multi}                & 99.8 & 98.3 & 99.3 & 98.6 & 83.3 & 98.4 & 98.0 & 98.3 & 85.8 & 98.3 & 90.5 & 99.3 & 98.9 & 96.6 & 98.8 & 86.3 & 99.8 & 87.3 & 99.8 & 96.1 & 95.6 \\
				
				CPCL~\cite{huang2022cpsd}                & 99.6 & 98.6 & 98.5 & 98.8 & 81.9 & 95.1 & 97.8 & 98.2 & 83.0 & 95.5 & 85.5 & 98.4 & 98.5 & 97.0 & 99.0 & 86.6 & 97.0 & 84.9 & 99.1 & 94.3 & 94.4 \\
				SST~\cite{chen2022sst}                   & 99.8 & 98.6 & 98.9 & 85.5 & 94.7 & 97.9 & 98.6 & 83.0 & 96.8 & 85.7 & 98.8 & 98.8 & 98.9 & 95.7 & 99.1 & 85.4 & 96.2 & 84.3 & 99.1 & 95.0 & 94.5 \\  
				\hline	
				\rowcolor{iyellow}Ours(SADCL)$^\star$      &100.0 & 99.0 & 99.5 & 99.1 & 88.9 & 98.8 & 98.7 & 99.6 & 84.2 & 98.4 & 90.1 & 99.4 & 99.6 & 99.0 & 99.3 & 90.2 & 99.6 & 88.9 & 99.8 & 95.3 & 96.4 \\ \hline 
			\end{tabular}
		}
		\caption{
			Comparisons with state-of-the-art methods on the VOC 2007 dataset. \emph{*} indicates the reproduced results of our implementation, and
			$\star$ indicates the model pre-trained on COCO.
			All of the inputs are 448$\times$448 resolution except SSGRL(576), ADDGCN(512) and KGGR(576).
		}
		\label{table: VOC2007 benchmark}
	}
\end{table*}

\begin{table*}[ht]
	\centering{
		 \setlength\tabcolsep{3pt}
		\resizebox{\textwidth}{!}{
			\begin{tabular}{c|c|c|c|c|c|c|c|c|c|c|c|c|c|c|c|c|c|c|c|c||c}
				\hline
				Methods & aero & bike & bird & boat &
				bottle & bus & car & cat & chair & 
				cow & table & dog & horse & motor & 
				person & plant & sheep & sofa & train & tv &
				mAP \\ \hline \hline
				VeryDeep~\cite{simonyan2015very}&99.1 &88.7 &95.7 &93.9 &73.1 &92.1 &84.8 &97.7 &79.1 &90.7 &83.2 &97.3 &96.2 &94.3 &96.9 &63.4 &93.2 &74.6 &97.3 &87.9 &89.0\\		        
				Fev+Lv~\cite{yang2016exploit}   &98.4 &92.8 &93.4 &90.7 &74.9 &93.2 &90.2 &96.1 &78.2 &89.8 &80.6 &95.7 &96.1 &95.3 &97.5 &73.1 &91.2 &75.4 &97.0 &88.2 &89.4\\
				HCP~\cite{wei2015hcp}           &99.1 &92.8 &97.4 &94.4 &79.9 &93.6 &89.8 &98.2 &78.2 &94.9 &79.8 &97.8 &97.0 &93.8 &96.4 &74.3 &94.7 &71.9 &96.7 &88.6 &90.5\\
				MCAR~\cite{gao2021learning}     &99.6  &97.1 &98.3 &96.6 &87.0 &95.5 &94.4 &98.8&87.0 &96.9 &85.0 &98.7&98.3&97.3 &99.0 &83.8 &96.8&83.7&98.3 &93.5  &94.3 \\
				SSGRL$^\star$~\cite{Chen2019ICCV}  & 99.7 &  96.1  & 97.7 &  96.5 &  86.9  & 95.8 &  95.0 &  98.9  & 88.3  & 97.6 &  87.4 &  99.1  & 99.2 &  97.3  & 99.0 &  84.8  & 98.3  & 85.8 &  99.2 &  94.1  & 94.8 \\
				ADDGCN$^\star$~\cite{ye2020attention}  & 99.8 &  97.1 &  98.6 &  96.8 &  89.4 &  97.1 &  96.5 &  99.3 &  89.0  & 97.7  & 87.5 &  99.2  & 99.1  & 97.7 &  99.1 &  86.3 &  98.8 &  87.0 &  99.3 &  95.4  & 95.5 \\
				KGGR$^\star$~\cite{chen2022knowledge}  & 99.8 & 97.3 & 98.4 & 97.1 & 87.9 & 97.3 & 96.5 & 99.3 & 89.4 & 97.8 & 88.7 & 99.4 & 99.4 & 97.9 & 99.2 & 86.3 & 98.8 & 86.3 & 99.7 & 95.2 & 95.6 \\
				DSDL~\cite{zhou2021dsdl}        &99.4 &95.3 &97.6 &95.7 &83.5 &94.8 &93.9 &98.5 &85.7 &94.5 &83.8 &98.4 &97.7 &95.9 &98.5 &80.6 &95.7 &82.3 &98.2 &93.2 &93.2 \\
				CCD-R101$^\star$~\cite{liu2022contextual}& 99.8 & 98.2 & 98.3 & 98.0 & 88.6 & 97.4 & 96.9 & 99.1 & 90.8 & 98.9 & 90.2 & 99.2 & 99.6 & 98.4 & 99.0 & 87.7 & 98.4 & 88.8 & 99.7 & 96.4 & 96.1 \\
				\hline
				\rowcolor{iyellow}Ours(SADCL)$^\star$ &99.9  &98.7  &99.0  &98.7  &91.2  &97.8  &97.4    &99.6  &92.3  &98.9  &90.3  &99.7  &99.7  &98.4  &99.4  &89.8  &99.1  &90.9  &99.7  &97.2  &96.9 \\
				\hline
			\end{tabular}

		}
		\caption{Comparisons with state-of-the-art methods on the VOC 2012 dataset. 
			$\star$ indicates the model pre-trained on COCO.
			All of the inputs are 448$\times$448 resolution except SSGRL(576), ADDGCN(512) and KGGR(576).}
		\label{table: VOC2012 benchmark}
	}
\end{table*}

\paragraph{Results on NUS-WIDE.}
NUS-WIDE is a web image dataset, with all images coming from Flickr.
The experimental results is reported in Table~\ref{table: nuswide benchmark}. Compared with the existing state-of-the-art methods, our method achieves significant performance on five evaluation metrics with 65.9\% mAP, 63.0\% CF1(All), 75.0\% OF1(All), 57.8\% CF1 (Top 3), and 70.6\% OF1 (Top 3). NUS-WIDE's image quality (resolution) is lower than MS-COCO's, so the recognition performance on NUS-WIDE of all methods has yet to reach the level of MS-COCO.
\begin{table}[t]
	\centering{
		\resizebox{\linewidth}{!}{
		\begin{tabular}{c|c| cc | cc }
			\hline
			\multirow{2}{*}{Methods}  & \multirow{2}{*}{mAP} & \multicolumn{2}{c|}{All}  & \multicolumn{2}{c}{Top 3}
			\\
			&         & CF1  & OF1   & CF1    & OF1  \\ \hline \hline
			CNN-RNN ~\cite{wang2016cnn}                     & -       & -    & -     & 34.7   & 55.2 \\
			ResNet101\textbf{$^*$} ~\cite{He2016CVPR}       & 62.5    & 59.2 & 73.8  & 54.6   & 69.4 \\
			CADM~\cite{chen2019cadm}                        & 62.8    & 60.7 & 74.1  & 56.3   & 70.6 \\
			ADDGCN\textbf{$^*$} ~\cite{ye2020attention}     & 63.3    & 60.3 & 73.5  & 56.5   & 69.3 \\
			P-GCN~\cite{chen2021learning}                   & 62.8    & 60.4 & 73.4  & 57.0   & 69.1 \\
			TDRG\textbf{$^*$} ~\cite{Zhao2021ICCV}          & 63.5    & 60.0 & 73.8  & 56.1   & 69.5 \\
			SST ~\cite{chen2022sst}                         & 63.5    & 59.6 & 73.2  & 55.9   & 68.8 \\ 
			MulCon~\cite{dao2021multi}                      & 63.9    & 61.8 & 74.8  & -      & -   \\
			CCD-R101\textbf{$^*$}~\cite{liu2022contextual}  & 64.2    & 61.8 & 74.6  & 56.7   & 70.0 \\ \hline
			\rowcolor{iyellow}Ours(SADCL)                     & \textbf{65.9}    & \textbf{63.0} & \textbf{75.0}  & \textbf{57.8}   & \textbf{70.6} \\ \hline
		\end{tabular}
	}
	}
	\caption{Comparisons with state-of-the-art methods on the NUS-WIDE dataset. \emph{*} indicates the reproduced results of our implementation. The input images are resized to 448$\times$448 resolution in both the training and testing phases.}
	\label{table: nuswide benchmark}
\end{table}

\paragraph{Results on Visual Genome.}
Visual Genome is a dataset recently used to evaluate multi-label image classification.
Following SSGRL~\cite{Chen2019ICCV} setting, we select the 500 most frequently occurring categories to build a new dataset called VG500.
Compared to other related datasets, this one presents more challenges due to the large number of categories.
In addition to the existing reports, ADDGCN and TDRG are reported using the corresponding official codes.
As can be seen from Table~\ref{table: vg500 benchmark}, we present five important evaluation metrics, including mAP, CF1(All), OP1(All), CF1(Top3), and OP1(Top3).
Except for CF1(Top3), our method outperforms other state-of-the-art methods, especially in mAP, obtaining 40.5\%.
\begin{table}
	\centering{
		\resizebox{\linewidth}{!}{
		\begin{tabular}{c|c|c| cc | cc }
			\hline
			\multirow{2}{*}{Methods} & \multirow{2}{*}{Resolution} & \multirow{2}{*}{mAP} & \multicolumn{2}{c|}{All}  & \multicolumn{2}{c}{Top 3}
			\\
			&                     &                & CF1  &  OF1 & CF1  & OF1  \\ \hline \hline
			ResNet101\textbf{$^*$}& 448$\times$448 & 35.3 & 29.3 & 47.0 & 16.0 & 27.8 \\
			ML-GCN                & 448$\times$448 & 32.6 & 27.5 & 42.8 & 16.8 & 27.1 \\
			ADDGCN\textbf{$^*$}  & 448$\times$448 & 37.3 & 33.2 & 47.8 & 19.8 & 28.4 \\  \hline\hline
			SSGRL                 & 576$\times$576 & 36.6 &  -   &  -   &   -  & -    \\
			ADDGCN\textbf{$^*$}  & 576$\times$576 & 38.2 & 33.5 & 47.8 & 19.9 & 28.4 \\
			KGGR                  & 576$\times$576 & 37.4 & 32.5 & 47.2 & 19.4 & 28.1 \\ 
			TDRG\textbf{$^*$}     & 576$\times$576 & 37.7 & 30.9 & 48.0 & 17.5 & 28.5 \\
			C-Tran                & 576$\times$576 & 38.4 & 35.2 & 49.5 & \textbf{20.1} & 28.7 \\
			MulCon                & 576$\times$576 & 38.5 &  -   &  -   &  -   &  -   \\
			Query2Label           & 576$\times$576 & 39.5 &  -   &  -   &  -   &  -   \\ \hline
			\rowcolor{iyellow}Ours(SADCL)       & 576$\times$576 & \textbf{40.5} & \textbf{41.0} & \textbf{54.0} & 17.1 & \textbf{29.2} \\ \hline
		\end{tabular}
		}
	}
	\caption{Comparisons with state-of-the-art methods on the VG500 dataset. \emph{*} indicates the reproduced results of our implementation.}
	\label{table: vg500 benchmark}
\end{table}

\subsection{Ablation Study}
In this section, we evaluate the effectiveness of each component of our proposed framework by performing an ablation study on MS-COCO and NUS-WIDE datasets in Table~\ref{tab:ablation}.
The results show that, on both datasets, employing only SARL significantly improves performance compared to the baseline. This improvement is from 81.5\% mAP to 85.0\% mAP and from 62.5\% mAP to 65.3.0\% mAP, respectively.
This demonstrates that SARL can learn category-related representations by focusing on category-related semantic regions.
We observe that SARL* drops 0.3\% mAP on both datasets compared to SARL, which is attributed to the Transformer-Encoder to model long-range dependencies from multiple objects and scene.
The collaborative learning of SARL and SSCL improves by 0.4\% mAP on MS-COCO and 0.3\% mAP on NUS-WIDE compared to SARL, demonstrating that SSCL can learn a better representation of potential features.
Combining the SARL and PSCL components, 85.4\% mAP and 65.7\% mAP were obtained on MS-COCO and NUS-WIDE. This indicates that PSCL can narrow the distance between positive label-level visual representations samples ($y_{ij}\!=\!1$) and category prototypes and increase the distance between negative class samples ($y_{ij}\!=\!0$) and category prototypes. In the end, when all components are used, the performance of our model is further improved, reaching 85.6\% mAP and 65.9\% mAP.
\begin{table}[t]
	\centering{
		\setlength\tabcolsep{2.5 pt}
		\renewcommand{\arraystretch}{1.25}
		\resizebox{\linewidth}{!}{
		\begin{tabular}{ccccc| ccc | ccc }
			\hline
			\multicolumn{5}{c|}{Module} & \multicolumn{3}{c|}{MS-COCO} & \multicolumn{3}{c}{NUS-WIDE} \\
			\cline{1-11}
			~ Baseline & SARL*      & SARL       & SSCL       & PSCL       & mAP & OF1 & CF1 & mAP & OF1 & CF1 \\
			\hline\hline
			\checkmark &            &            &            &            & 81.5 & 79.7 & 76.0 & 62.5 & 73.8 & 59.2 \\
			\checkmark & \checkmark &            &            &            & 84.7 & 81.5 & 78.7 & 65.0 & 74.2 & 61.0 \\
			\checkmark &            & \checkmark &            &            & 85.0 & 81.7 & 79.3 & 65.3 & 74.8 & 61.7 \\
			\checkmark &            & \checkmark & \checkmark &            & 85.4 & 81.7 & 79.6 & 65.7 & 74.9 & 62.7 \\
			\checkmark &            & \checkmark &            & \checkmark & 85.4 & 82.1 & 79.8 & 65.6 & 75.0 & 62.4 \\
			\checkmark &            & \checkmark & \checkmark & \checkmark & \textbf{85.6} &  \textbf{82.1}  & \textbf{79.8} & \textbf{65.9} & \textbf{75.0} & \textbf{63.0} \\
			\hline
		\end{tabular}
	}
		\caption{Ablation study of different components in MS-COCO and NUS-WIDE Datasets.
			Baseline mean ResNet101 network with average pooling and fc layer.
			Noteworthy, when Baseline is combined with additional components, the pool and fc layers in Baseline will be removed.
			SARL* denotes SARL after removing Transformer-encoder. All metrics reflect all predicted scores instead of taking the top-3 highest prediction scores.}
		
		\label{tab:ablation}
	}
\end{table}

\vspace{-0.4cm}
\subsection{Visualization Study}
\begin{figure}[b]
	\centering
	\subfigbottomskip=-5pt
	\subfigure{
		\hspace{-1.5mm}
		\begin{minipage}{0.5\textwidth}
			\includegraphics[width=86mm,height=43mm]{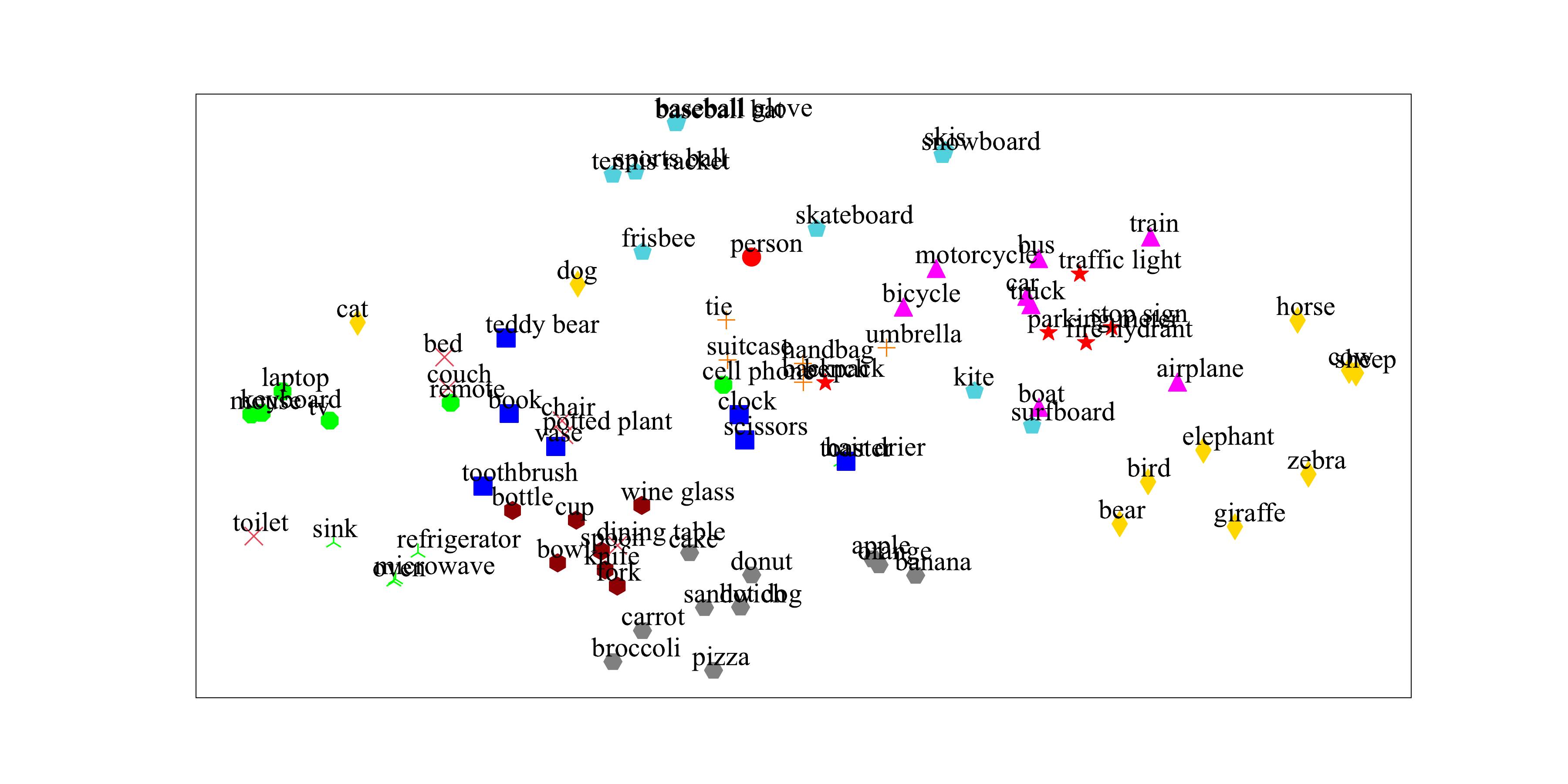}
	\end{minipage}}
	
	\subfigure{
		\hspace{-1.5mm}
		\begin{minipage}{0.5\textwidth}
			\includegraphics[width=86mm,height=7mm]{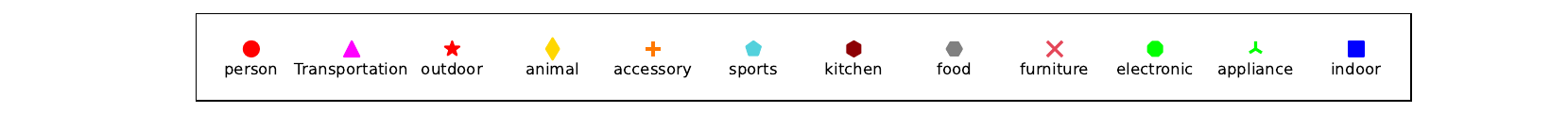}
	\end{minipage}}
	\caption{Visualization of the learned category prototypes on the MS-COCO dataset. Different colors and shapes mean different superclasses.}
	\label{label_prototype}
\end{figure}

In order to further explore category prototypes, we use t-SNE~\cite{tsne2008visualizing} to visualize them as illustrated in Figure~\ref{label_prototype}.
It is clear to show that our category prototypes contain meaningful semantic information.
Specifically, the distance between different categories is as large as possible, and paired categories with high co-occurrence have a smaller distance than paired classes with low co-occurrence.
Therefore, the correlation between the labels is well modeled by our method, and the category prototypes are discriminative.

\begin{figure}[t]
	\centering
	\includegraphics[width=1\columnwidth]{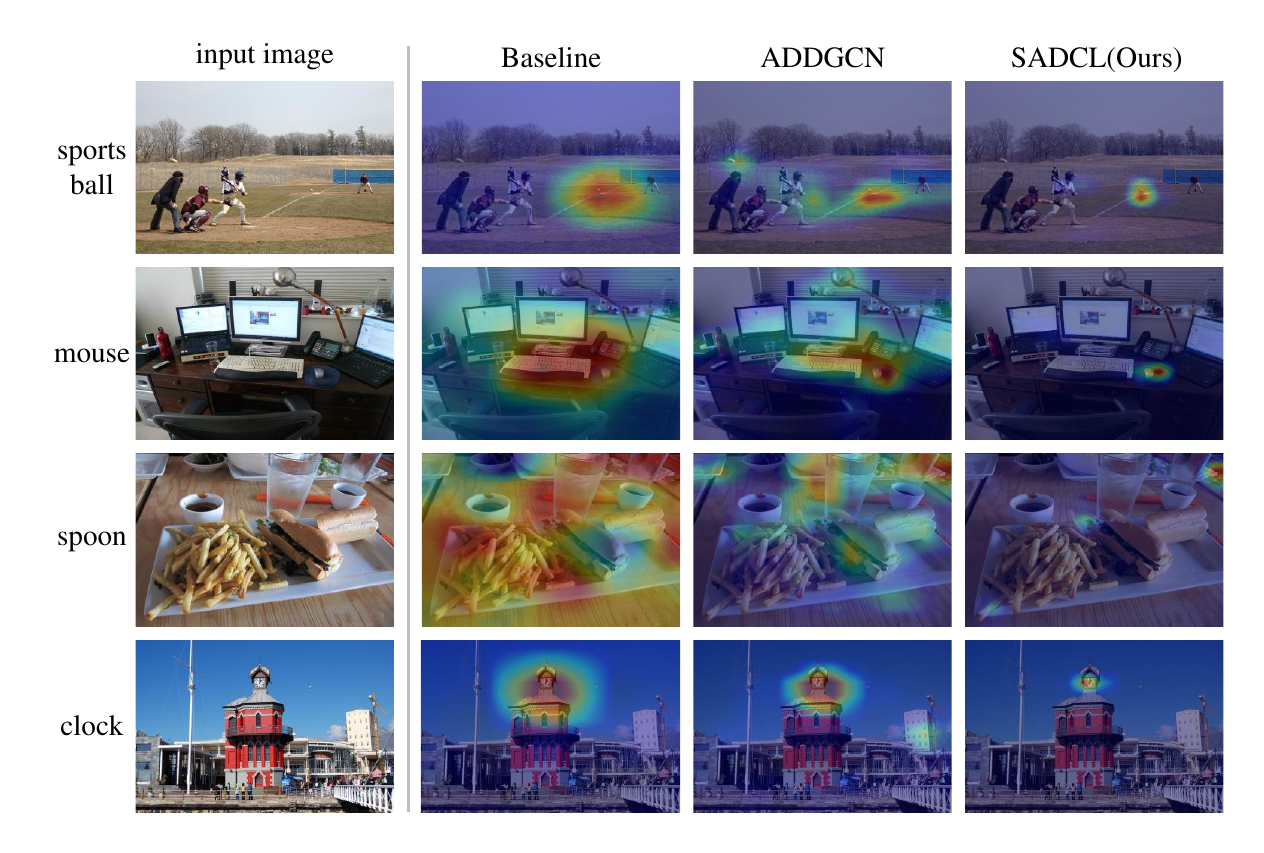}
	\caption{Visualization analysis of baseline method, ADDGCN method, and our method SADCL.}
	\label{semantic_map}
\end{figure}

To evaluate the performance of our proposed model for localizing discriminative object regions, we visualize category-related attention maps for SARL modules and category-related activation maps for baseline and ADDGCN.
The results from Figure~\ref{semantic_map} show that the baseline method and the ADDGCN method localize ambiguously and easily generate noise. In contrast, our method identifies the most discriminative part of the related object.	
As shown in the 3th row of Figure~\ref{semantic_map}, from the \textit{spoon} category, we can observe: i) the baseline method fails to localize related regions, ii) the ADDGCN method can focus on \textit{spoon} while generates a lot of noise, and iii) our method can accurately localize discriminative regions for \textit{spoon}.

\begin{figure}
	\centering
	\subfigbottomskip=-5pt
	\subfigure{
		\hspace{-1.5mm}
		\begin{minipage}{0.5\textwidth}
			\includegraphics[width=86mm,height=43mm]{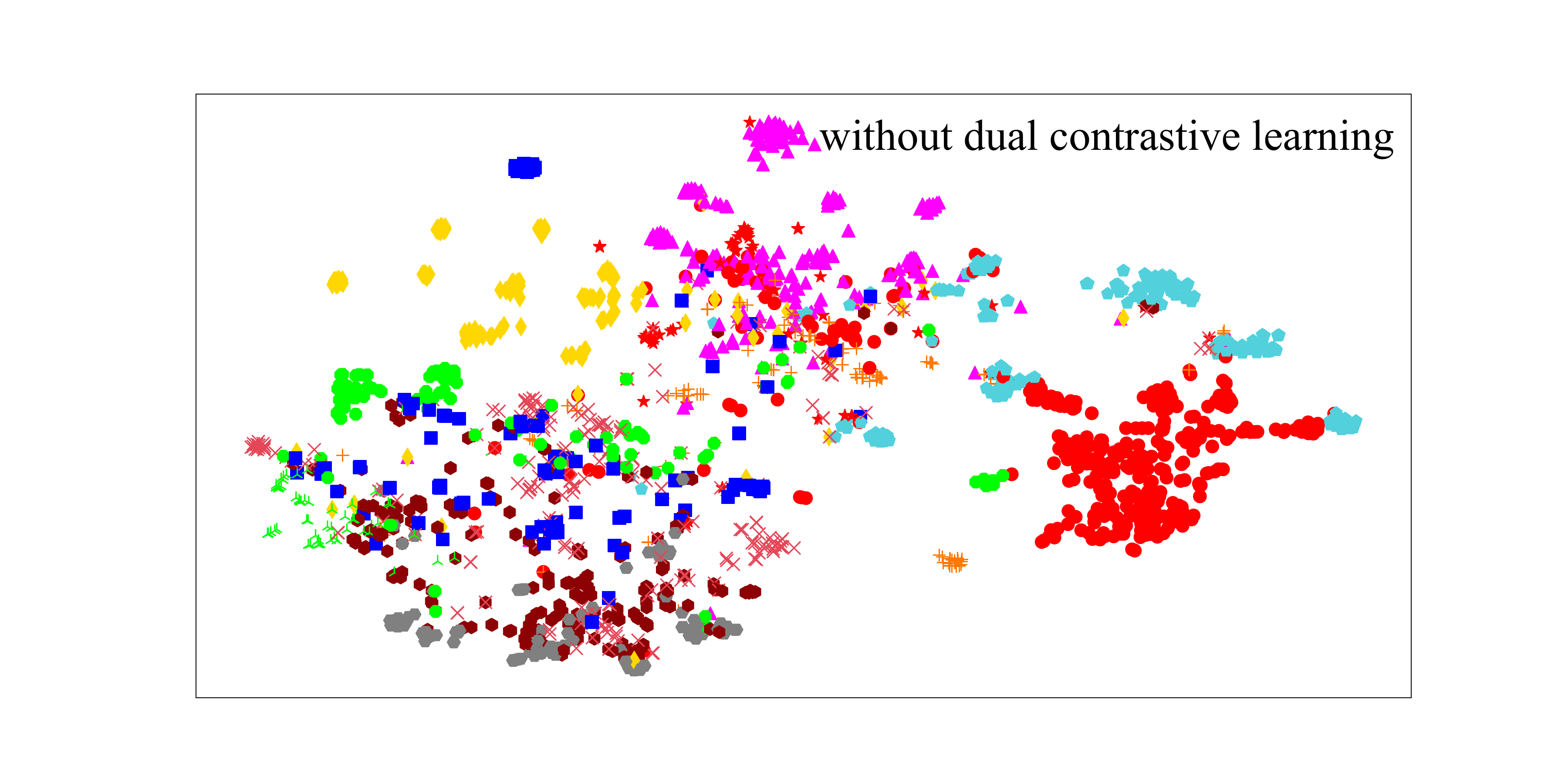}
	\end{minipage}}
	
	\subfigure{
		\hspace{-1.5mm}
		\begin{minipage}{0.5\textwidth}
			\includegraphics[width=86mm,height=43mm]{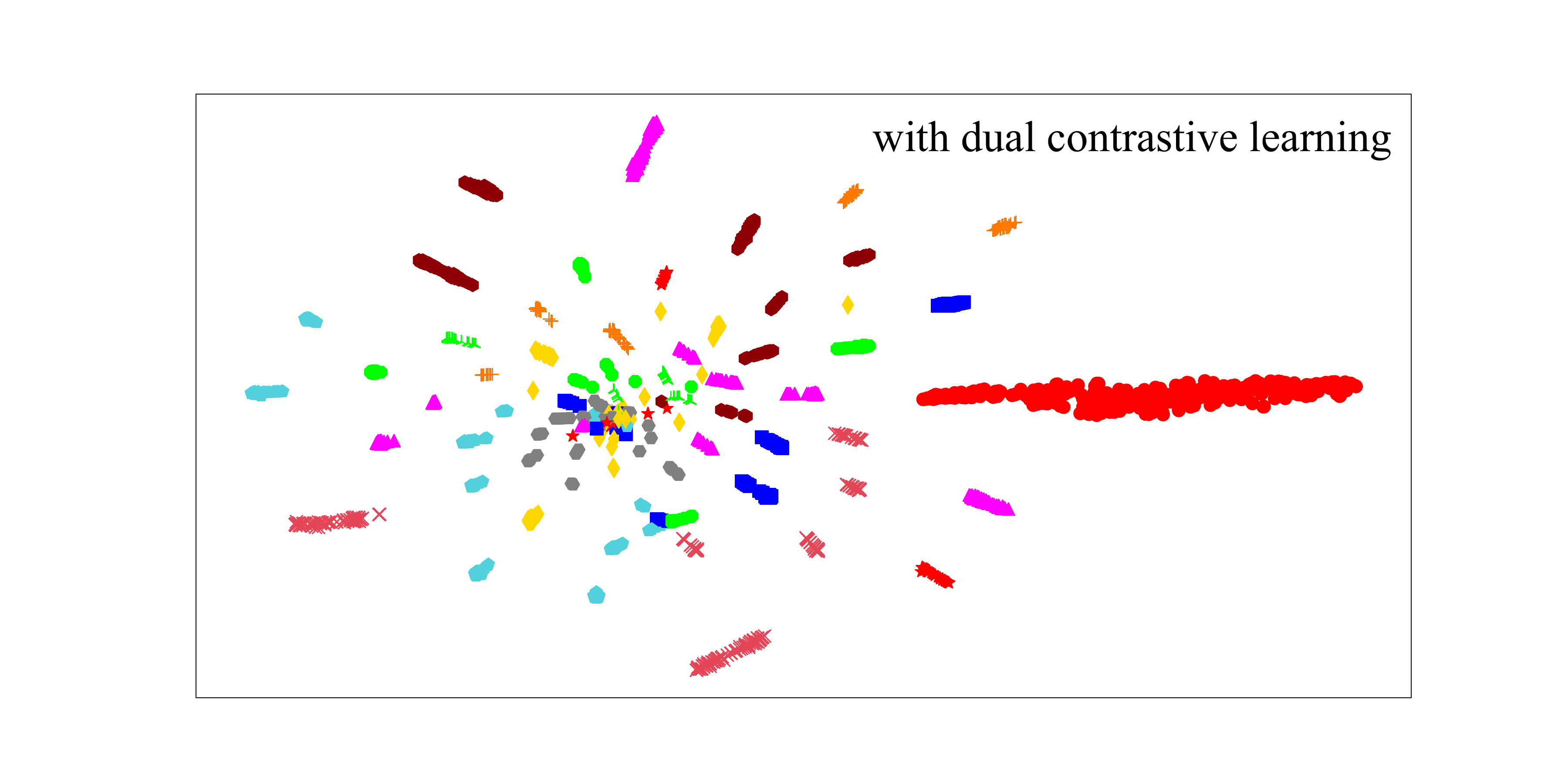}
	\end{minipage}}
	
	\subfigure{
		\hspace{-1.5mm}
		\begin{minipage}{0.5\textwidth}
			\includegraphics[width=86mm,height=7mm]{imgs/word.pdf}
	\end{minipage}}
	
	\caption{Visualization of the 2000 learned label-level visual representations randomly sampled images of the MS-COCO test dataset.}
	\label{learned_features}
\end{figure}

In order to evaluate the quality of the learned representations, we show the t-SNE visualization of 2000 activated label-level visual representations ($y_{ij}\!=\!1$). In a total of 12 superclasses, the samples of most classes are discretely scattered in the original space. As shown in Figure~\ref{learned_features}, each class representation is clustered in its own relatively independent region after using our dual contrastive learning strategy.

\section{Conclusion}
In this paper, we propose a novel semantic-aware dual contrastive learning model with semantic-aware representation learning, sample-to-sample contrastive learning, and prototype-to-sample contrastive learning.
By intervening in SARL, our proposed method focuses on local regions of related categories and learns label-level visual representations.
The collaborative learning of SSCL and PSCL can model complex intra-category and inter-category relationships so that the label-level feature representation with positive labels can be close to the category prototype.
Experiments on five popular benchmarks demonstrate the effectiveness and state-of-the-art performance of our method.

\ack This work was supported in part by the National Natural Science Foundation of China (No.61876002, 62076005, U20A20398), 
the Natural Science Foundation of Anhui Province (No.2008085UD07, 2008085MF191), 
the University Synergy Innovation Program of Anhui Province, China (No.GXXT-2021-002, GXXT-2021-030, GXXT-2021-065, GXXT-2022-029), 
the National Natural Science Foundation of China and Anhui Provincial Key Research and Development Project(No.202104a07020029).


\bibliography{ecai}
\end{document}